\documentclass[Afour,sagev,times]{sagej}
\usepackage{algorithmicx}
\usepackage{algorithm}
\usepackage{algpseudocode}
\usepackage{moreverb,url}
\usepackage{textcomp}
\usepackage{tabularx,ragged2e}
\newcommand{\RNum}[1]{\uppercase\expandafter{\romannumeral #1\relax}}
\usepackage[colorlinks,bookmarksopen,bookmarksnumbered,citecolor=red,urlcolor=red]{hyperref}

\newcommand\BibTeX{{\rmfamily B\kern-.05em \textsc{i\kern-.025em b}\kern-.08em
T\kern-.1667em\lower.7ex\hbox{E}\kern-.125emX}}

\def\volumeyear{2016}

\begin{document}

%\runninghead{Smith and Wittkopf}

\title{PLeDO: Pain Level Detection for Osteoarthritis from EMR Data}

\author{Yuhao Chen\affilnum{1}, Jiahao Cai\affilnum{1}, Nafiz Sadman\affilnum{1}, Farhana Zulkernine\affilnum{1}, John Queenan\affilnum{2}, and David Barber\affilnum{2}}
%\author{Alistair Smith\affilnum{1} and Hendrik Wittkopf\affilnum{2}}

\affiliation{\affilnum{1}School of Computing, Queen's University, Kingston, Ontario, Canada \\
\affilnum{2}Department of Family Medicine, Queen's University, Kingston, Ontario, Canada}

\corrauth{Yuhao Chen, School of Computing, Queen's University, Kingston, Ontario, Canada}

\email{yuhao.chen@queensu.ca}

\begin{abstract}
Osteoarthritis (OA) is a progressive chronic joint disease resulting in a breakdown of articular cartilage and bone when damaged joint tissues are not able to normally repair themselves. The aim of this pilot research study is to understand the pain severity for OA from patients' primary care Electronic Medical Records (EMR), both from the structured medical data and the unstructured chart note data using information extraction, natural language processing and machine learning techniques. We propose SPaDe, a Synonym-based Pain level Detection tool to categorize patients into having mild or moderate-to-severe pain to understand diagnosis and treatment methods based on only the pain related expressions in the unstructured chart note. Expressions are subjective, objective, and influenced by cultural background and demography which poses a difficult challenge. Therefore, we improve the model by incorporating the medication information from the structured EMR data and pain scale related information from the chart note to propose an integrated pain level detection tool for OA called PLeDO. With the help of human labeled gold standard data, we demonstrate that both SPaDe and PLeDO can detect mild and moderate-to-severe pain from the EMR data to analyze and potentially improve the quality of care in primary care setting.
\end{abstract}


\maketitle

\section{Introduction}
\label{sec:introduction}
Prevalence and incidence rates for OA are problematic to establish due to variations in diagnostic definitions \cite{b10}. Arthritis Alliance of Canada estimated that there may be over 4.4 million people living with OA in Canada and in 30 years this number can reach 10 million (1 in 4 Canadians) \cite{b1}. With a growing aging  population in Canada and increasing rates of obesity and inactivity, the rate of OA is projected to increase from 13.8\% to 18.6\% between 2010 and 2031 \cite{b3}.

Comorbidity is common in OA population and approximately 59 - 87\% of people with OA have at least one other chronic illness \cite{b4}. Approximately 80\% of the individuals with OA exhibit some degree of movement limitations, and 25\% are not able to perform their regular daily activities of life \cite{b5}. Pain associated with OA and its severity significantly impacts the health‐related quality of life (HRQOL) and productivity of the affected population \cite{b6}. The loss in costs due to productivity or work associated with OA in Canada is substantial, which was estimated to be \$12 billion in 2010 and to reach \$17.5 billion Canadian dollars in 2031 \cite{b3}. Pain is the most disabling symptom of OA and a major driver of clinical decision making and heath care utilization \cite{b17}. Therefore, identifying symptomatic OA patients with moderate to severe pain in a real-world setting would be integral to understanding the burden of the disease and the treatment journey. Information about the pain medication can help pharmaceutical companies to innovate new treatment options.

The Canadian Primary Care Sentinel Surveillance Network (CPCSSN) is a multi-disease EMR surveillance system \cite{b22}. It consists of over 2 million patients' data collected from 1,500 participating primary care clinicians. CPCSSN's data extraction algorithm only extracts the structured data items such as date of birth, gender, and disease from EMR, but not the unstructured chart note data, which often contains valuable clinical information. Traditional manual extraction/audit of unstructured and free text data from clinical notes and other narratives is time consuming, labor intensive and expensive. Information Extraction (IE) can relieve some of these problems by enabling automated extraction of the relevant information into a research database and thereby, facilitating further analysis of the data using a combination of Natural Language Processing (NLP) and Machine Learning (ML) techniques \cite{b28, b29, b30}. NLP facilitates the processing of semi-structured and unstructured text data in narrative clinical documents to identify and label medical terminology, and interpret written information \cite{b31} \cite{b52}. It also helps to extract and transform text data into numeric vectors that can be passed to ML algorithms for prediction and decision support \cite{b32, b33, b34, b58}.  \\
\indent Currently there are no disease-modifying agents available in the market for OA. Non-pharmacologic and pharmacologic therapies used for management of OA are aimed at improving pain, disability, and quality of life \cite{b4}. The most common pharmacological treatment options for the symptomatic treatment of OA are acetaminophen, NSAIDs (topical/oral), Cox2 inhibitors, duloxetine, intra-articular (IA) corticosteroids, IA hyaluronic acid, tramadol or strong opioids \cite{b12}\hspace{1sp}\cite{b13}. OA embodies one of the most frequently occurring painful conditions \cite{b14}. Pathophysiology of OA pain is complex, exhibiting a combination of nociceptive and neuropathic mechanisms involved in both the local and central levels \cite{b14}. 

In Canada, OA pain information is unlikely to be systematically documented in the EMR. To the best of our knowledge, no studies in Canada have yet attempted to identify and characterize patients with moderate-to-severe OA pain based on EMR chart note data in the primary case setting. The use of longitudinal EMR data is very useful for surveillance of a population that is at high risk of or is diagnosed with OA \cite{b20}. CPCSSN \cite{b21}\cite{b22}\cite{b24} has developed and validated an algorithm that identifies patients diagnosed with OA in Canadian primary care using readily available structured data extracted from patients’ EMR \cite{b23}. However, this algorithm currently cannot classify OA pain levels into mild, moderate-to-severe based on the descriptive chart note data. 

\textbf{Contribution:} In this research, we explore and develop a variety of information extraction (IE) methods using NLP and machine learning (ML) methods to retrieve important information from the CPCSSN EMR structured and  unstructured chart note data. We fabricate these methods into multiple IE pipelines to integrate and encode the data to train a classification model. The aim is to determine the severity of pain experienced by OA patients mainly based on the unstructured text data in the physicians' notes and also the structured medication data. 
The contributions of this research are as follows.

\begin{itemize}
    \item To the best of our knowledge, this is the first study that utilizes integrated NLP and ML techniques to identify pain levels in patients with Osteoarthritis using physicians' chart note data and structured medication information from EMRs.
    \item We present an unsupervised synonym-based clustering approach, SPaDe, which extracts and clusters pain related expressions from the unstructured chart note data to categorize patients into mild and moderate-to-severe and validate it using gold standard manually labeled data.
    \item We propose a novel integrated Pain-level Detection tool for OA, PLeDO, which combines (a) SPaDe, the pain expression-based clustering approach, (b) a pain scale-based approach, and (c) a medication-based approach to detect OA pain level using both structured medication and unstructured chart note data. PLeDO categorizes patients into mild and moderate-to-severe pain categories for studying treatment patterns in the primary care setting. 
    \item We provide a scalable, less resource-intensive methodology for pain level detection, contributing to advancements in clinical NLP. The framework can inspire further research into unsupervised methods for other healthcare text classification tasks, expanding the toolkit available to the community.
    \item An ablation study is presented to demonstrate the improvements achieved at different stages when extending SPaDe with the additional information to build PLeDO. 
\end{itemize}

The rest of the paper is organized as follows. We describe the related work in Section \ref{sec:relwork}. The methodology is discussed in Section \ref{sec:methodology} which provides an overview of the analytic workflow and explains the study sample.  Section \ref{sec:implementation} illustrates the implementation details about SPaDe, the pain scale and the medication based approaches to pain level categorization. The experimental results are presented in Section \ref{sec:valres} with discussions about the observations and outcomes. Finally, Section \ref{sec:conclusion} concludes the paper with a list of future work directions.
\section{Related Work}
\label{sec:relwork}
\subsection{Background}
%\subsection{Pain Categorization}
%OARSI/OMERACT initiative\cite{b15} categorizes the patient-reported pain experience for knee and hip OA in the context of disease progression as follows.
Patient reported pain experiences for knee and hip OA in the context of disease progression was categorized by OARSI/OMERACT initiative as follows \cite{b15}.
\begin{itemize}
\item Early OA: Predictable, sharp or other pain brought on by specific triggers, eventually limiting high impact activities but affecting little the other low impact activities. 
\item Mid OA: Predictable, more constant pain occurring in association with joint symptoms (e.g., joint locking). This pain affects daily activities such as walking. 
\item Advanced OA: Pattern of constant dull and aching pain with intermittent unpredictable episodes of intense pain which leaves a person exhausted. This results in a significant avoidance of social and recreational activities \cite{b15}.
\end{itemize}
  
Various patient-reported outcomes have been used to assess pain and disability in hip and knee OA \cite{b8} \cite{b18}. One of the most widely used tools is the Western Ontario and McMaster Universities OA Index (WOMAC) \cite{b16}, which consists of three sub-scales: pain, stiffness, and physical function. For the evaluation of OA pain, Visual Analog Scale (VAS) or Numerical Rating Scale (NRS) assessment of pain intensity are also commonly used \cite{b7}. Patients may be asked about the experience of “pain, aching of stiffness in or around the knee” over a specific time frame.

\subsection{Pain Related Classification}

In recent years, a growing number of studies have explored the intersection of pain assessment and machine learning to improve pain management and tailored treatment plans \cite{b49,b50,b51}. Most existing literature focuses primarily on numerical data, with limited exploration of text data using supervised learning. Lotsch et al. \cite{b45} incorporated supervised machine learning to analyze numerical data from preoperative cold pain tests to predict persistent pain after breast cancer surgery. High negative predictive values (94\%) were achieved, though positive predictive values were low (10\%). Results highlight the role of the endogenous pain inhibitory system in pain persistence. Similarly, Alambo et al. \cite{b48} explored the use of statistical machine learning methods, such as logistic regression and decision trees, to distinguish patterns between patients with and without pain. Their best-performing model achieved a remarkable 98\% F1 score, demonstrating the potential of these methods for accurate pain classification.

Beyond numerical data, several studies have investigated the use of clinical text for pain-related classification. DiMartino et al. \cite{b46} evaluated the feasibility of using NLP to detect uncontrolled symptoms (moderate or severe pain) in clinical notes from 1,644 hospital encounters for cancer patients. They used the machine learning models in Clinical Annotation Research Kit to do classification and achieved 61\% accuracy and 69.5\% F1 score. Their findings demonstrated the initial feasibility of NLP in identifying symptom burden but emphasized the need for further development before such tools can be reliably implemented in clinical workflows. In another study, the authors in \cite{b47} analyzed 235,789 clinical texts from an Emergency Department Information System using BlueBERT, a domain-specific BERT model pre-trained on PubMed abstracts. The model was employed for binary classification (pain or no pain), achieving an impressive 95\% accuracy on the evaluation set.

Recent studies also indicate that pain- and symptom-related phenotyping from clinical narratives remains feasible but highly task-specific. In a directly relevant recent study, Hughes et al. \cite{b54} investigated pain-related classification in emergency-department clinical text, showing that large-scale unstructured notes can support automated pain-focused modeling in acute-care settings. More recently, note-based pain and symptom analytics have expanded into adjacent supervised clinical-NLP tasks. For example, a hybrid machine-learning and LLM framework was reported to predict short-horizon cancer pain episodes from combined structured and unstructured EHR data \cite{b55}. Oncology note-based studies \cite{b56} have also demonstrated the feasibility of extracting symptom presence and severity directly from treatment notes using BERT-family models. Related recent work in psychiatry notes further suggests that even in contemporary clinical NLP, carefully designed rule-based systems remain competitive with or superior to larger language models when the corpus is small, expert-annotated, and clinically specialized \cite{b57}.

Taken together, these studies confirm the value of numerical data, structured clinical variables, and narrative EMR text for pain- and symptom-related phenotyping. While supervised learning approaches demonstrate promise, obtaining expert annotations is often prohibitively expensive and time-consuming. Although clustering methods have been applied to pain level identification in prior research \cite{b49}, these efforts have predominantly focused on numerical data, which is less complex and challenging compared to unstructured textual data.

Existing studies are predominantly supervised and depend on manually labeled datasets, disease-specific annotations, or infrastructure that supports large pretrained models. In contrast, our work focuses on osteoarthritis pain severity identification from unstructured primary-care EMR notes. Gold-standard labels were unavailable at model-development time and the workflow had to operate in an air-gapped environment because of sensitive patient data. Accordingly, our contribution is distinct from recent supervised note-classification studies: we address OA pain stratification in a no-label, privacy-constrained setting using an unsupervised framework tailored to real-world primary-care EMR data. The challenge motivated us to explore and develop more scalable and cost-effective approaches for leveraging clinical text in healthcare research.
\section{Methodology}
\label{sec:methodology}
In this section, we discuss the key challenges that influenced the algorithms, a brief overview of our approaches, creation of the study sample, and the study environment. 

\subsection{Challenges}
\label{sec:challenges}

Unstructured text data analytics offer many challenges.
\begin{itemize}
    \item Length of the sentences are not uniform.
    \item Text data can contain spelling errors, domain specific terminology, ill structured data with missing ``,", ``.", ``'", and incorrect grammar and phrases.
    \item Text data is often written in hybrid format with numeric data, dates, acronyms, emoji, and other form of literal expressions.
    \item Text data can also contain expressions written in multiple languages in the same document.
    \item Each chart note often contains duplicate data from previous notes because the new chart entry for a patient's visit is created by first copying the text from the previous note and then appending new information to it.
\end{itemize}

The specific type of medical chart note data that we had in this study, offered the first three challenges more frequently. The expressions, as mentioned in the $4^{th}$ point, were not in different languages but had wide variations for pain levels. Therefore, extraction of the words posed a serious challenge for the classification of the medical records or patients with mild, moderate, or severe pain. 

Each patient's EMR is identified by an ID which helps to link personal data such as age, demography, and location, to multiple other data such as the chart notes and medications. However, a patient can have many years of notes and medications. Therefore, extracting and linearizing patients' records especially those with many years of data can be a big challenge. When feeding this linearized data to machine learning models, the size disparity of the data creates further problems as some of the data are too long while others may have just one encounter record.

Another problem with processing multiple years of data is that a patient can have different records containing expressions of varying pain levels. After treatment, pain can reduce and then increase again. How should such patients be categorized? Based on extensive discussion with the medical experts and collaborators, we decided about the following policy for classification. 

\begin{itemize}
    \item If a patient has multiple notes expressing different pain levels, label the patient with the highest pain level expression. 
    \item As expressions of pain can vary for different culture, race, and ethnicity, synonyms should be considered to represent each pain level instead of specific words.
    \item Only pain related medications should be considered for this OA pain related study to understand the treatment pattern. 
\end{itemize}  
  
An overview of our methodology is presented next followed by a description of the data used in this study.
%For each patient, we compiled the medication information from all the records of that patient and filtered out non-pain related medications.

%We are going to discuss the selection of the study sample for this study in Section \ref{sec:study_sample}.

\subsection{Overview of PLeDO}
\label{sec:overview}

The key objective of this study is to develop novel NLP and ML techniques to categorize OA patients based on their reported pain levels into mild and moderate-to-severe categories in the primary care EMR data. We propose an integrated Pain Level Detection (PLeDO) tool to categorize OA patients into mild and moderate-to-severe pain groups given their structured  EMR data containing medications and unstructured chart note data. We use the medication information from the structured parts of the EMR data. The unstructured chart notes contain valuable information such as scale of pain level as logged by the primary care physician during the patient encounter. The notes also contain information shared by the patients about their pain which can vary widely based on patients' demography, culture, age, and gender. We aim to extract such expressions including the pain scale information and the medication information and use the same in developing the integrated system, PLeDO, which consists of the following 3 subsystems.
\begin{enumerate}
    \item Pain Expression-based system - Extracts patients' expressions regarding pain levels from the chart note data by developing 3 word dictionaries representing 3 pain levels: mild, moderate, and severe. A word embedding method is also used to transform the words to vectors to facilitate word similarity calculation. The similarity of each word in patients' notes from each dictionary is computed to classify notes and hence the patients into mild or moderate-to-severe category.
    \item Pain Scale-based system - Extracts pain scale related information recorded by some physicians in the unstructured chart note data.
    \item Medication-based system - Extracts and compiles prescribed medications, referral to pain clinics, surgical treatments, or use of assistive tools from all records spanning multiple years from the structured EMR data for each patient. In parallel, Pain related medications are grouped into mild, and moderate-to-severe pain levels. Based on the types of medications prescribed to them, patients are categorized to the highest level pain category. 
\end{enumerate}

The workflow of the algorithm is given in Figure \ref{fig:overall_flowchart}. Further details about each subsystem are given in Section \ref{sec:methodology}.

\subsection{Study Sample}
\label{sec:study_sample}

\begin{figure}[h]
    \centering
    \includegraphics[width=3.5 in]{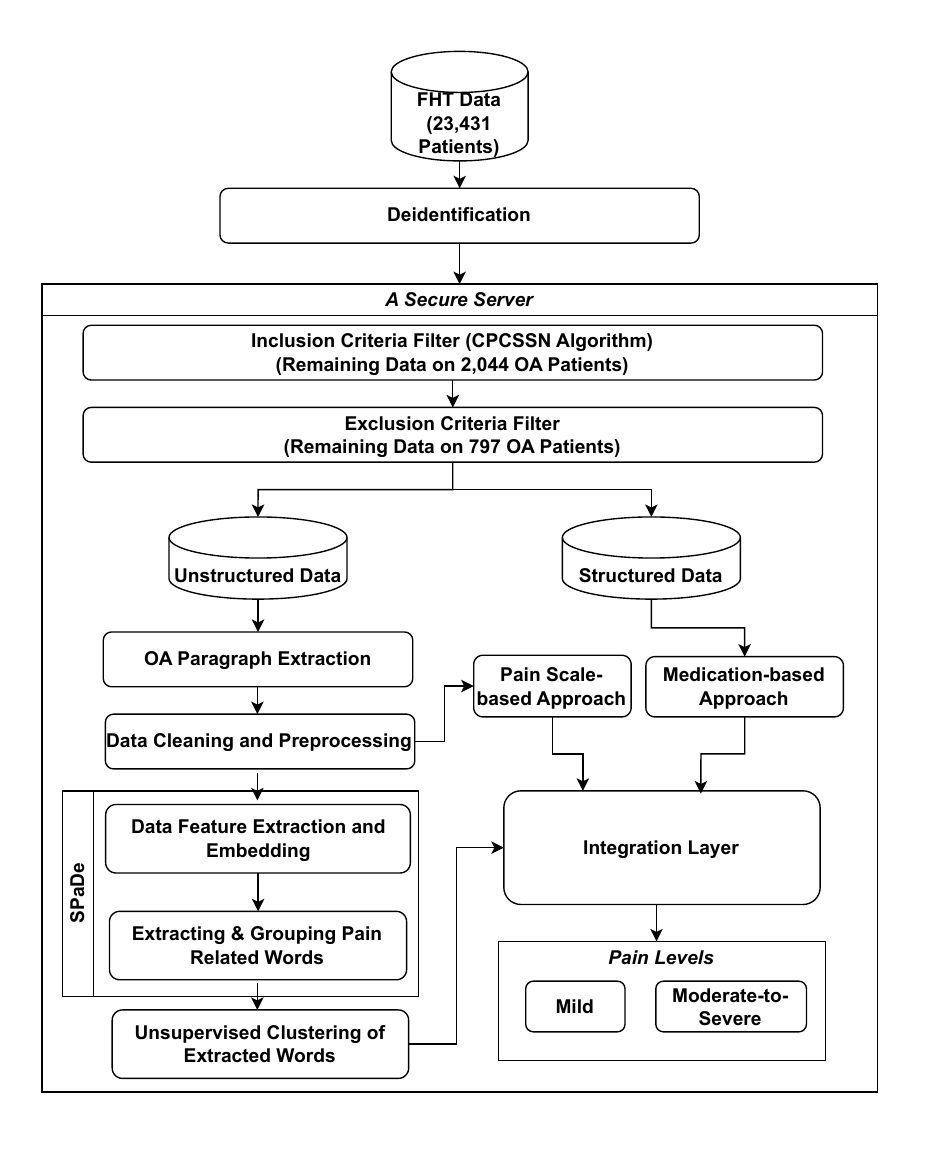}
    \caption{The Overall Workflow}
    \label{fig:overall_flowchart}
\end{figure}

The patient sample included in this study is extracted from OSCAR \cite{b25} open source EMR system used by the primary case physicians participating in this study. The sample data includes primary care patient population having a diagnosis of OA based on the CPCSSN case definition algorithm. Figure \ref{fig:overall_flowchart} shows the overall workflow of this research. The complete process involved multiple steps. First, the CPCSSN data extraction algorithm customized for the various EMR systems was used to extract and store selected data items on a secured staging server, where the data is deidentified using the TiDE \cite{b35}\cite{b44} by the CPCSSN data scientists. %In a clinical note, TiDE \cite{b44} replaces Protected Health Information (PHI) \cite{b35} with surrogate values. Users have the ability to configure TiDE and choose which data categories to de-identify. In order to apply a time offset to dates and ages, users can also declare one. TiDE uses a number of methods to find and de-identify PHI. Named Entity Recognition (NER) and pattern matching using regular expressions are the primary methods. An attacker or unauthorized person with access to the data would not be able to distinguish between real values and synthetic replacements if PHI were replaced with plausible surrogate values. The remainder of the text hides the PHI that was missed. 
Then the deidentified data was shared with our research group on another virtual machine customized for secure data analysis. Since the chart note data often contains names and Protected Health Information (PHI), all PHI needed to be removed or deidentified, and the research ethics approval needed to be in place before the data could be shared with us. 

We created an initial study sample based on the following inclusion criteria. Then an exclusion criteria was applied to filter out specific types of patients' records to prepare the final study sample.

 %\begin{figure}[h]
 %     \centering
  %    \includegraphics[width=2.8in,height=2.5in]{OA_Population.jpg}
   %   \caption{Subject Population Distribution}
    %  \label{fig:oa_subj_dist}
  %\end{figure}
 
\subsubsection{Inclusion Criteria} 
\label{sec:inclusion}

CPCSSN regularly collects and integrates the structured EMR data into the database but the unstructured chart notes are not included in the integrated database. Therefore, for this study we had to follow the data request procedure to extract the unstructured chart notes from the EMR systems of a regional family health team (FHT) of practitioners. All registered patients of FHT of practitioners are by default considered to have given consent to such medical studies unless they explicitly opt out via an established procedure. 

A total of 23,431 patients were registered in the EMR systems of the selected regional FHT whose data were extracted for this study to create the initial patient population. Out of these patients, only 2,044 patients of all ages were identified as having OA based on the CPCSSN OA case definition algorithm \cite{b23}. In a separate study, CPCSSN researchers published an algorithm to identify and label OA patients with ICD-9 disease codes based on specific criteria and the available data fields in the EMR \cite{b23}. We used that label as the gold standard label to identify and include OA patients in our study sample. We included only the OA patients' EMR data for the period of 12 years from 2010-2022. We further narrowed down the dataset by applying the following exclusion criteria to focus on specific type of OA related pain for the study purposes. 

\begin{table}[h]
\caption{Sample Size Description}
\small\addtolength{\tabcolsep}{-5pt}
\begin{tabular}{|l|c|}
\hline
\multicolumn{1}{|c|}{\textbf{Creation of Study Sample}}                    & \multicolumn{1}{l|}{\textbf{Num. of Patients}} \\ \hline
Total OA Patients (2010-2022)                                   & 23,431                                             \\ \hline
After applying Inclusion Criteria                                  & 2,044                                             \\ \hline
After applying Inclusion and Exclusion Criteria                                  & 797                                             \\ \hline
Total Samples to Create Gold Standard Data                                      & 269                                              \\ \hline
Gold Standard Samples Missing Pain Info & 113                                              \\ \hline
Final Gold Standard Samples                            & 156                                              \\ \hline
\end{tabular}
\end{table}

\subsubsection{Exclusion Criteria}
\label{sec:exclusion}

From the OA patient population consisting of 2,044 patients, we excluded all patients having the following ICD criteria (and all ICD-9-CM subcodes).

\begin{enumerate}
    \item Inflammatory joint diseases such as
    \begin{enumerate}
        \item Rheumatoid arthritis (ICD-9 code: 714)
        \item Psoriatic arthritis (ICD-9 code: 696)
        \item Ankylosing spondylitis (ICD-9 code: 720)
        \item Septic arthritis (ICD code: 711)
        \item Erythematous conditions (including lupus) (ICD code: 695)
        \item Gout (including pseudogout) (ICD code: 274)
    \end{enumerate}
    \item Systemic metabolic bone disease (e.g., Crystal arthropathies ICD-9: 712, Paget’s disease or Osteitis deformans without mention of bone tumor, ICD-9 code: 731; Disorders of mineral metabolism such as metastatic calcifications ICD-9 code 275)
    \item Other disorders of soft tissues such as Fibromyalgia and Neuropathic pain, ICD-9 code:729)
\end{enumerate}

Following the application of exclusion criteria, we identified a total of 797 OA patients. We then obtained the deidentified EMR structured and unstructured data in multiple CSV files. The unstructured chart note data was processed to extract pain related expressions and information regarding the use of pain scales by the physicians to measure pain levels. A subset of this data was manually reviewed to create the gold standard data. The structured data was used for gender based analysis and mainly to obtain pain related medications prescribed to the patients.

\subsubsection{Patient Demography}

We extracted additional structured data such as patients' gender and age to report about the distribution of patients in our study population. %based on a) gender, b) age, c) Body Mass Index (BMI), and d) comorbidity (other diseases that the patient was diagnosed with in the past) suffering from severe or moderate pain.

\subsubsection{Gold Standard Data} \label{sec:golddata}

We randomly selected a subset of the OA patients from our study sample for manual labeling with mild, moderate, and severe pain levels. The sample size was calculated based on 26\% prevalence (patients with severe pain) with an expected sensitivity of 80\% and specificity of 80\% at 95\% confidence interval and 10\% precision. We considered the manually labeled data as the gold standard data for the development and validation of our algorithms to classify OA pain levels. The manual chart validation was conducted by a person with years of experience and formal training in labeling medical data. To reduce the bias and ensure the consistency, the annotator was guided by predefined protocols developed in collaboration with medical experts. Additional, he was also supported by other medical expert collaborators who provided guidance during the annotation process. In the case of ambiguity or confusion, other experts were consulted to ensure accuracy of the labels. A total of 269 OA patients out of the 797 patients in the study sample were selected randomly for manual evaluation. Based on the results of evaluation, among these 269 patients, chart notes for 156 patients included mentions of their pain levels, while the remaining 113 patients did not have any explicit reference to pain levels in their records. Therefore, for the purpose of evaluating our algorithm, we utilized the data of the 156 patients only which had pain level information explicitly mentioned in their chart notes.

\subsection{Study Environment}
\label{sec:environment}

For this study, the deidentified data was staged on a secured data analytics environment called the Restricted Data Environment (RDEN) for access, analysis, and reporting. For the $1^{st}$ objective of categorizing pain level, we developed an information extraction and transformation pipeline using NLP techniques to process the unstructured text data in the EMR chart notes. The overall data processing includes deidentification, information extraction and cleaning, and developing approaches to predict pain levels as mild, or moderate-to-severe based on the EMR chart note data. We combined ``moderate" and ``severe" into one category because the treatments heavily overlap for these two categories. However, if needed, the categories, moderate and severe, can be separated with a negative effect on the accuracy of each subcategory as the algorithms fail to determine the category correctly solely based on medication information.

\section{Implementation} 
\label{sec:implementation}

Our integrated system (PLeDO) comprises 3 different approaches for categorizing the pain levels namely pain expression-based (SPaDe), pain scale-based, and medication-based approaches. We first explain the general data pre-processing methods that applies to one or more of our proposed approaches. The implementation of each of the approaches are described subsequently under specific subsections.

\subsection{General Data Pre-processing}
\label{sec:data_process}

We built the information extraction (IE) algorithms based on a manual visual inspection of the data to identify the key challenges in processing the data, positions of the desired information in the data, layout of the data, and other cues such as context, keywords, or symbols that can be used to develop the IE algorithms. The key challenges with processing EMR data were discussed under Section \ref{sec:challenges}. Accordingly, to achieve the objectives of pain level classification, we built custom information extraction, cleaning, preprocessing, and transformation algorithms considering the structure, layout, context, and representation of the EMR structured and unstructured data. The extracted information was used for data linking and analysis, and transformed into different representations to feed into machine learning algorithms. 

\subsubsection{OA Paragraph Extraction:}
\label{sec:para_extract}

As explained under Section \ref{sec:challenges}, each chart note often contains  text from the previous chart notes and includes details about the complete historical record of the patient with multiple diseases. Consequently, extraction of pain related information returns a lot of noise from different diseases and not just OA. Therefore, we conducted a keyword search method to extract paragraphs related to OA from the unstructured chart note data. We constructed a comprehensive keyword dictionary containing keywords associated with OA. Subsequently, we searched these keywords in each chart note. If any of the keywords were identified in a note, the sentence containing the keywords, as well as two preceding and two succeeding sentences were extracted. Finally, for each patient, we combined all the OA related extracted contents from all EMRs into a combined note. 

\subsubsection{Data Cleaning and Preprocessing:}
The EMR chart note and transformed deidentified structured data in the CSV files required robust data cleaning and formatting for information extraction. Characters such as ``," were removed from medication and demographic information. As mentioned earlier, often previous notes are copied and then new notes are appended when logging the chart note data creating much duplication in the data. Eliminating duplicate data was necessary to accurately calculate frequencies from the data. Similarly, numerical data may not be germane to the analysis and can cause problems during data processing. For example, a patient's height, weight, BMI, and other numeric information such as time, and date are not useful for the analysis. We applied Natural Language Toolkit (NLTK) \cite{b26} methods to remove the special characters, and punctuations. We developed algorithms to compare and remove duplicate records, and create a cleaner version of the data.

\subsubsection{Data Feature Extraction:}

To extract all potentially significant data features such as verbs, adverbs, and adjectives from the EMR notes, we applied the NLTK \cite{b26} Named Entity Recognition (NER) and Parts of Speech (POS) taggers to identify and tag the significant terms for this study. Sentiments such as (positive sentiment) ``improved", ``feeling better", (negative sentiment) ``worsening" are indicative of a patient's emotional states with a focus on pain management. A regular sentiment analysis method cannot address the complex and subjective nature of expression and experiences of the patients focused on pain. Therefore, we use NER and POS to identify and tag words which are often used to describe pain such as verbs: ``ache”, ``hurt”, ``burn”; adjectives: ``sharp”, ``dull”; adverbs: ``intensely”, and ``slightly”. 

After tagging, we developed algorithms to extract the words from the annotated chart note data for the different approaches.

We applied regular expression based data extraction for the pain scale based approach as explained in the respective section. For expressions such as date or pain scale where the data has a specific format, regular expressions can be used to effectively extract such data.

\subsubsection{Word Embedding:}

Word embedding \cite{b27} converts words into numerical vectors in a high-dimensional space to be processed efficiently by computer algorithms. Older approaches such as one hot encoding or TF-IDF \cite{b43} can not capture the context information. Machine learning based models \cite{b27} are able to process contextual information and create similar embeddings for synonymous words. We converted the extracted word features to vectors using GLOVE embedding \cite{b27}, which gave good performance and was computationally cost-effective compared to Med-BERT \cite{b41} more complex embedding methods. Since the GloVe embeddings rely on a pretrained vocabulary, any words with typos is removed in the embedding generation process. The nature of our framework, which relies heavily on similarity comparisons between each word and each pain-level embedding. The pretrained language models such as BERT-style models are not inherently designed for direct similarity comparisons at the word level or sentence level, resulting in suboptimal performance for our specific use case \cite{b53}. While models like Sentence-BERT \cite{b53} address some limitations in similarity comparison, they are pre-trained and fine-tuned on general domain data. In contrast, the GloVe embedding version we used was trained on a substantial amount of medical-specific vocabulary, making it particularly suitable for our dataset and task requirements. This domain-specific vocabulary played a crucial role in improving the accuracy of our framework.
\begin{figure}[h]
    \centering
    \includegraphics[width=3.5in]{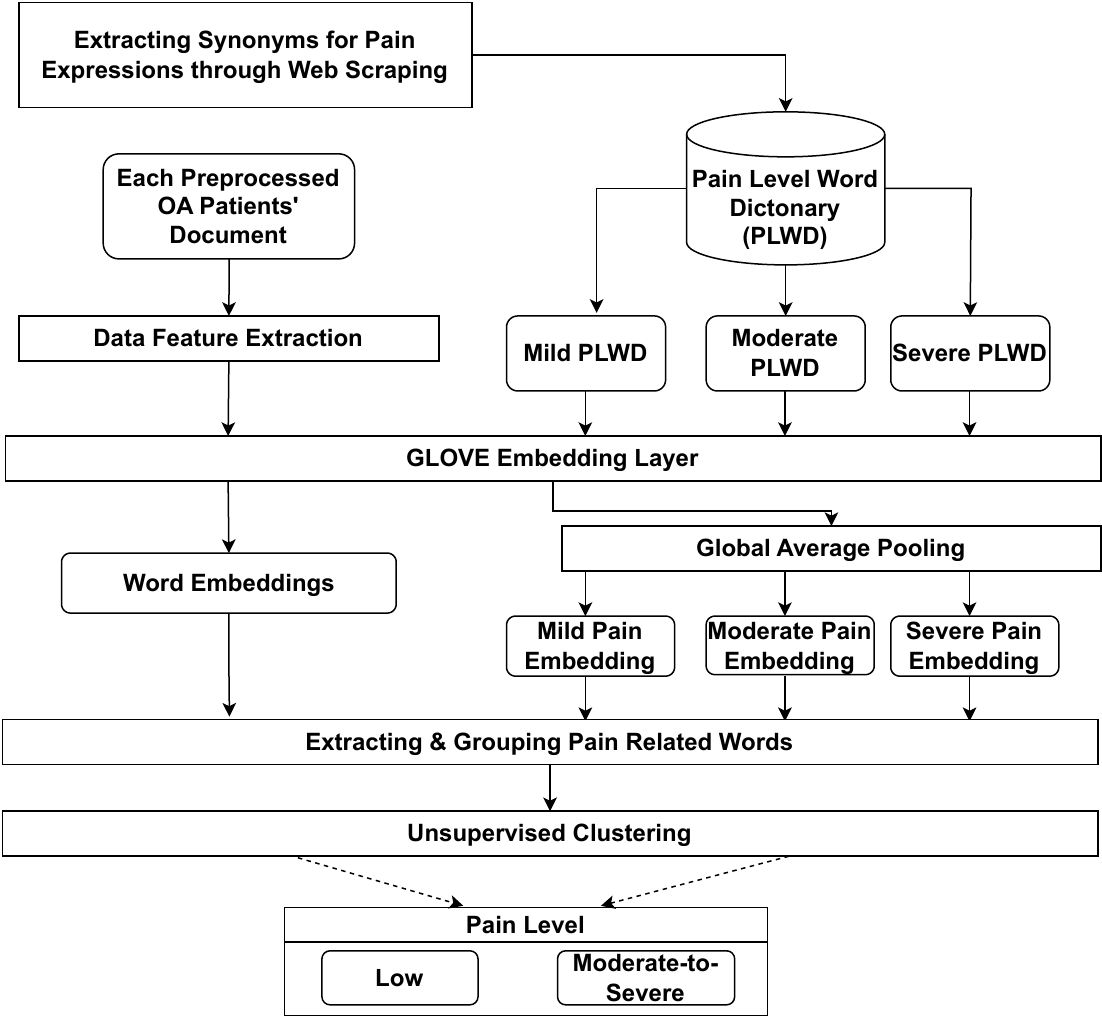}
    \caption{The Workflow of Pain-Level Clustering}
    \label{fig:pain_level_workflow}
\end{figure}
\subsection{Approach \RNum{1}: Pain Expression-based Approach}
\label{sec:syn_app3}

Manually labeling all chart notes for supervised learning was not feasible especially because the size of a single note varied from 1 to 9,488 words and there were 482,185 notes to label. Therefore, we developed an unsupervised learning approach to cluster the notes into mild, and moderate-to-severe categories based on words expressing different pain levels. Then we used the gold standard data to validate our approach. 

Patients use subjective, situational, cultural and emotional words to describe their pain level based on their background, demography, race, and gender. Therefore, we defined 3 Pain Level Word Dictionaries (PLWD), each representing a collection of words or synonyms aligned with one of the 3 pain levels, namely mild, moderate, and severe. 
%The expressions in the chart notes can be related to other health problems than OA.

To extract relevant expressions for OA, we first extracted OA related sentences and then applied different NLP techniques to clean and process the extracted data as explained under Section \ref{sec:data_process}. Next we applied GLOVE embedding to convert the text to numeric vectors. Figure \ref{fig:pain_level_workflow} depicts the different processing steps. The rest of this section explains the different steps in more detail such as creating the PLWDs, using them to group and extract pain related words from the chart notes, and applying unsupervised clustering to determine the centroids of the extracted word clusters. Finally, the similarity of the combined chart note of each patient, represented by the average value of the extracted word embeddings, to the centroids is used to categorize the note or the patient under either mild and moderate-to-severe pain category.

\subsubsection{Pain Level Word Dictionary:}

We developed three Pain Level Word Dictionaries (PLWD) by compiling three distinct sets of English vocabularies with 50 synonyms in each dictionary indicating mild, moderate, or severe pain level. We used Beautiful Soup\footnote{Beautiful Soup: https://tedboy.github.io/bs4\_doc/}, a web scraping library, to extract data from a specific website\footnote{Website: https://www.thesaurus.com/} providing a digital thesaurus and a tool for identifying synonyms. After creating these dictionaries automatically, we filtered the results by ChatGPT \cite{b37}\hspace{1sp}\cite{b38} to eliminate irrelevant or redundant terms and thereby, ensure that the final set of synonyms accurately described the corresponding pain level. Specifically, ChatGPT was used to evaluate whether each candidate word appropriately described pain intensity corresponding to the intended severity category. Words that did not clearly reflect pain severity or that were redundant were flagged for removal. Importantly, all filtered results were manually reviewed by a medical expert before finalizing the dictionaries. The medical expert examined each retained term to ensure that it accurately described osteoarthritis-related pain severity in a clinical context. Only those terms deemed clinically appropriate were included in the final PLWDs.The final PLWD contained 32, 31, and 37 synonyms indicating mild, moderate and severe pain levels. Examples of synonyms are shown in Table. \ref{tab:plwd}. 

%\begin{figure}[!h]
%    \centering   %\includegraphics[width=3.5 in]{keywords.png}
%    \caption{Examples of words from the 3 PLWDs indicating the three pain levels.}
%    \label{fig:keyword}
%\end{figure}}

\begin{table}[!h]
\caption{Examples of words from the 3 PLWDs indicating the three pain levels}
\footnotesize
\begin{tabular}{|c|ccccc|}
\hline
PLWD     & \multicolumn{5}{c|}{Example words from the dictionary}                                                                                      \\ \hline
Mild     & \multicolumn{1}{c|}{low}     & \multicolumn{1}{c|}{slight}    & \multicolumn{1}{c|}{mild}        & \multicolumn{1}{c|}{small}    & tingling \\ \hline
Moderate & \multicolumn{1}{c|}{aching}  & \multicolumn{1}{c|}{tearing}   & \multicolumn{1}{c|}{hurting}     & \multicolumn{1}{c|}{medium}   & sharp    \\ \hline
Severe   & \multicolumn{1}{c|}{beating} & \multicolumn{1}{c|}{grounding} & \multicolumn{1}{c|}{lancinating} & \multicolumn{1}{c|}{crushing} & heavy    \\ \hline
\end{tabular}
    \label{tab:plwd}
\end{table}
Words in the PLWDs were used as keywords to search for and extract semantically similar words from the chart notes indicating the three pain levels.  

\subsubsection{Computing Centroids of PLWDs:}
\label{alg:centroids_PLWD}

The PLWDs are instrumental for the SPaDe algorithm to categorize the chart notes to different pain levels. We applied the word embedding method to convert all the words in the PLWDs into numeric vectors and then calculated the centroids ($L^{centroid}, M^{centroid}, H^{centroid}$) as the average of these vectors for each PLWD as illustrated in Algorithm \ref{alg:centroid}. In the algorithm, L, M, and H correspond to 'Low', 'Moderate', and 'High', denoting the set of words in text format from mild, moderate, and severe pain categories respectively. In lines 2-6, the words are transformed into word vectors using GLOVE embedding. Lines 7-9 calculates the centroids. These centroids serve as the reference points in determining the proximity of words in the chart notes to each PLWD or pain level for categorizing the chart notes.

\begin{algorithm}
\caption{Centroid Detection for the PLWDs} \label{alg:centroid}
    \begin{algorithmic}[1]
        \Procedure{Centroid}{$L$,$M$,$H$}
        \State Initialize three empty array: $L^{glove}, L^{glove}, L^{glove}$
        \For{$L_k, M_k, H_k$ in $L, M, H$}      
            \State $L^{glove}.append(GLOVE(L_k))$
            \State $M^{glove}.append(GLOVE(M_k))$
            \State $H^{glove}.append(GLOVE(H_k))$
        \EndFor
        \State $L^{cent}= Average(L^{glove}, axis=0) $
        \State $M^{cent}= Average(M^{glove}, axis=0) $
        \State $H^{cent}= Average(H^{glove}, axis=0) $            
        \State \textbf{return} $L^{cent}, M^{cent},H^{cent}$
        \EndProcedure
    \end{algorithmic}
\end{algorithm}

\subsubsection{Extracting \& Grouping Pain Related Words:} 
\label{sec:grouping}

We developed an algorithm to search the combined chart note for synonyms or semantically similar words given the words in each PLWD, and extracted them to build mild, moderate, and severe word clusters for each patient as presented in Algorithm \ref{alg:word_gouping}. The similarity was calculated based on the proximity of a word vector to the centroids of the 3 PLWDs. The combined chart note refers to the OA related notes extracted from multiple visits, which were then combined, preprocessed, and embedded for each patient. The algorithm calculated the Cosine Similarity of each word vector $N_w$ in the processed note $N$ to the three centroids (mild, moderate, and severe) using Eq.\ref{equation} and assigned $N_w$ to the cluster of the closest centroid. A larger cosine similarity value indicates a stronger semantic relationship and the closest centroid. 
%We utilized metric (Equation \ref{equation}) to calculate the distance between $N_w$ and each pain level centroid.

\begin{equation} \label{equation}
    \text{similarity}(\mathbf{a},\mathbf{b}) = \frac{\mathbf{a} \cdot \mathbf{b}}{\|\mathbf{a}\| \|\mathbf{b}\|}
    = \frac{\sum_{i=1}^{n} a_i b_i}{\sqrt{\sum_{i=1}^{n} a_i^2} \sqrt{\sum_{i=1}^{n} b_i^2}},
\end{equation}

where $w$ represents $w^{th}$ word in the note, $\cdot$ denotes the dot product, $\|\mathbf{a}\|$ and $\|\mathbf{b}\|$ denote the Euclidean norms of vectors $\mathbf{a}$ and $\mathbf{b}$, respectively, and $n$ is the dimensionality of the vectors.

\begin{algorithm}
\caption{Synonym-based Pain Level Grouping Algorithm} \label{alg:word_gouping}
    \begin{algorithmic}[1]
        \Procedure{Grouping}{$N$,$L$,$M$,$H$}
        \State Initialize three cluster to 0: $C^{L},C^{M},C^{H}$
        \State $L^{centroid}, M^{centroid},H^{centroid} = Centroid(L, M, H)$
        \For{$N_w$ in $N$}   \label{code:distance1}
            \State $ S^{N_w, L^{centroid}} = Similarity(N_w , L^{centroid}) $ \label{code:distance2}
            \State $ S^{N_w, M^{centroid}} = Similarity(N_w , M^{centroid}) $\label{code:distance3}
            \State $ S^{N_w, H^{centroid}} = Similarity(N_w , H^{centroid}) $\label{code:distance4}
            \If{$S^{N_w, L^{centroid}} $ is the largest $\And$ $S^{N_w, L^{centroid}} > T^{similarity}$} \label{code:distance5}
                \State $C^{L}$ += 1 \label{code:distance6}
            \ElsIf{$S^{N_w, M^{centroid}} $ is the largest $\And$ $S^{N_w, M^{centroid}} > T^{similarity}$}\label{code:distance7} %here distance4 was used - NAFIZ
                \State $C^{M}$ += 1 \label{code:distance8}
            \ElsIf{$S^{N_w, H^{centroid}} > T^{similarity}$} \label{code:distance9}
                \State $C^{H}$ += 1\label{code:distance10}
            \EndIf \label{code:distance11}
        \EndFor \label{code:distance12}
        \State $TotalWords = C^{L} + C^{M} + C^{H}$
        \State $C^{L} =  \dfrac{C^{L}}{TotalWords}$
        \State $C^{M} =  \dfrac{C^{M}}{TotalWords}$
        \State $C^{H} =  \dfrac{C^{H}}{TotalWords}$
        \State \textbf{return} $C^{L}$, $C^{M}$, $C^{H}$
        \EndProcedure
    \end{algorithmic}
\end{algorithm}

Lines 3-14 in Algorithm \ref{alg:word_gouping} show the similarity computation and assignment of words to the cluster of the closest centroid. Here, $C^{L}$, $C^{M}$, and $C^{H}$ denote the mild, moderate, and severe cluster centroids respectively. A threshold value $T^{similarity}$ was used to filter out word vectors that are too far (very low Cosine Similarity) from all the centroids. Based on the experiment, we found that lower thresholds (e.g., \textless 0.6) allowed the inclusion of numerous irrelevant words, which introduced noise into the results. In contrast, higher thresholds (e.g., \textgreater 0.6) excluded a significant number of words, including some that were relevant but expressed with slight variations in phrasing or language. As a result, we selected a threshold value of 0.6. Word vectors were included only if the similarity was greater than a predefined threshold value of 0.6 since they did not provide any useful information or were not relevant to pain expressions. Lines 15-18 computes the distribution of the 3 categories of word vectors for each patient's combined and processed chart note data denoted by $N$. We observed that the mild pain level words occurred most frequently in most notes, followed by the moderate and then the severe pain-level words, respectively. It  indicates that mild pain levels were more commonly reported than higher pain levels.  %Finally, the cluster with the highest number of sentences in the document is selected as the winning cluster to represent the pain level. 

\subsubsection{Unsupervised Clustering of Extracted Words:}
\label{sec:unsupervised}

The large frequency of words indicating mild pain level created a problem in identifying the optimal thresholds of word frequency based on which notes could be categorized into mild, moderate, or severe pain levels. %Without ground truth labels, it is extremely challenging to determine the optimal threshold that accurately determines the percentage of high or moderate pain level words required to ascertain whether a patient belongs to high or moderate pain level category. 
Therefore, we developed an unsupervised clustering approach using the K-Means clustering algorithm \cite{b36} to determine the centroids of all the word vectors extracted from the whole study sample as shown in Algorithm \ref{alg:pain_detector}. We chose K-Means because our goal was to partition pain descriptions into stable, severity-consistent groups in a continuous embedding space and K-Means provides deterministic, interpretable centroids that reflect prototypical severity expressions. Since pain expressions in clinical notes are noisy and subjective, SPaDe cannot guarantee perfect severity inference for every expression. Using a highest-severity aggregation would make the method overly sensitive, as a single misclassified high-severity expression could dominate the patient-level label. To address this, we assign patient severity based on the most frequently inferred severity category across all chart notes, trading sensitivity for robustness and reducing the impact of occasional misclassifications. $Notes$ indicate all the notes of all patients in the study sample and $N_n$ represents each patient's note. Each note is transformed into a 3 dimensional vector representation using Algorithm \ref{alg:word_gouping}, where each dimension represents the percentage of mild, moderate, and severe word vectors respectively in that note. Then the set of all these 3D vectors from all notes in the study sample are passed on to a K-Means clustering algorithm which divides them into 3 clusters indicating mild, moderate, and severe pain.
\begin{algorithm}
\caption{Pain-Level Detector using K-Means Clustering} \label{alg:pain_detector}
    \begin{algorithmic}[1]
        \Procedure{Detector}{$Notes$,$L$,$M$,$H$}
        \State Initialize an empty array: $P$
        \For{$N_n$ in $Notes$} 
            \State $C^{L},C^{M},C^{H} = Grouping(N_n,L,M,H)$
            \State $P.append([C^{L},C^{M},C^{H}])$
        \EndFor
        \State $prediction = KMeans(P, n\_cluster=3)$
        \State \textbf{return} $prediction$
        \EndProcedure
    \end{algorithmic}
\end{algorithm}

The KMeans algorithm works by randomly selecting three initial centroids. Then, it iteratively assigns each data point to its nearest centroid. After each epoch, it updates the centroids by averaging the data points in each cluster. This process is repeated until the centroids converge, which means that the data points no longer change their assigned cluster. We conducted experiments with various randomly selected initial centroids, and the final outcomes were found to be similar. One example of final centroids learned by the K-Means clustering algorithm are shown in TABLE \ref{tbl:centroid}. 
\begin{table}[!h]
\caption{Centroid of KMeans Clustering Algorithm} 
\begin{center}
\begin{tabular}{|l|l|l|l|}
\hline
           & \textbf{Mild} & \textbf{Moderate} & \textbf{Severe} \\ \hline
\textbf{0} & 0.61         & 0.26         & 0.13          \\ \hline
\textbf{1} & 0.51         & 0.28         & 0.21          \\ \hline
\textbf{2} & 0.48         & 0.38         & 0.14          \\ \hline
\end{tabular} \label{tbl:centroid}
\end{center}
\end{table}

The rows in the result table show 3 centroids in the extracted word vector space. Values in the columns for each row indicate the average percentage occurrence of mild, moderate, and severe pain level words in the respective cluster. Row 0 shows the highest value for mild words in the first column. Thus it represents the centroid for mild category. Similarly, row 1 has the highest value in severe column and therefore, represents the severe category. Row 2 has the highest value in the moderate column and denotes the moderate category. Thus all notes (denoted by the transformed word vectors $C^L$, $C^M$, and $C^H$) in the respective clusters of centroids 0, 1, and 2 are categorized as mild, severe, and moderate pain level. 

%Finally, since the objective was to have two categories, mild and moderate-to-severe, we combined the notes under moderate and severe categories into moderate-to-severe category.

We decided a priori to focus on detecting individuals with moderate to severe pain. Our preliminary work on SPaDe showed that often similar words are used to describe moderate to severe pain and it is difficult to separate these two levels accurately. Pain is a subjective experience and different expressions are used to describe the pain in the chart note based on patients’ descriptions. Even human experts struggled to decide about moderate and severe pain levels from some of the chart notes. Therefore, we decided to develop the proof-of-concept tool from this pilot study to focus on two categories instead of three. Simplifying the analysis allowed us to streamline the process.

%This is accomplished by a sentence-level grouping which is able to find the minimum Euclidean distance between each sentence embedding and pain-level embedding.  A shorter Euclidean distance threshold indicates a stronger semantic relationship between the words. Finally, the cluster with the highest number of sentences in the document is selected as the winning cluster to represent the pain level. This grouping process is the key step in predicting pain levels based on the sentiment of a patient's EMR note, allowing for more personalized and effective pain management and treatment. The workflow of our NLP pipline is shown in Fig. \ref{fig:pain_level_workflow}.

%\begin{algorithm}
%\caption{Pain-Level Detector} \label{paindetection}
 %   \begin{algorithmic}[1]
  %      \Procedure{PainLevelDetection}{$Notes$,$L$,$M$,$H$}
   %     \STATE Initialize an empty array: $S$
    %    \FOR{$N_n$ in $Notes$} 
     %       \STATE $C^{L},C^{M},C^{H} = Grouping(N_n,L,M,H)$
      %      \STATE $TotalWords = C^{L} + C^{M} + C^{H}$
       %     \STATE $T^{H}, T^{M} = Threshold(Notes)$

        %    \IF{$C^{H} >= TotalWords * T^{H}$}
         %       \STATE $S.append(High)$
          %  \ELSIF{$C^{M} >= TotalWords * T^{M}$}
           %     \STATE $S.append(Moderate)$
           % \ELSE
           %     \STATE $S.append(Low)$
           % \ENDIF
        %\ENDFOR
       % \EndProcedure
   % \end{algorithmic}
%\end{algorithm}

\subsection{Approach \RNum{2}: Pain Scale-based Approach}
\label{sec:rule_app2}
We preprocessed the data to extract valid pain level entries in the “score/range” format (e.g., “6/10”) while excluding unrelated patterns such as dates. All the processing were performed on the same 797 study sample. 
%After a comprehensive analysis of the unstructured data, we found that the majority of pain level entries follow a specific format denoted as “score/range”. Given that the pain level ranges from 0 to 10, our initial step involved extracting all OA related notes that included the keyword “/10”. Then, we used a regular expression “[0-9]+/[0-9]+” to identify all the phrases with “score/range” pattern. However, we noticed that not all instances of this pattern referred to pain levels. Some were associated with dates, such as “09/10/2019”. To filter out this kind of noise, we implemented an additional filtering algorithm. This algorithm constrained the extraction of the ``range" to 10, paired with a ``score" either less than or equal to 10. Moreover, the scores prefixed with “0” were also removed.  
\begin{algorithm}
\caption{Pain Scale-based Categorization} \label{alg:rule_base}
    \begin{algorithmic}[1]
        \Procedure{painscale}{$N_n$}
        \State Initialize one empty array: $S$
        \State Initialize one array that contains severity related keywords: $K$
        \State $Score^{list}$ = re.findall($``[0-9]+/[0-9]"$, $N_n$) \label{regular_expression}
        \For{$Score^{list}_w$ in $Score^{list}$} 
            \State Extract the number before ``/" as $Score$
            \State Extract the number after ``/" as $Range$
            \If{$int(Range)$ == $10$ $\And$ $int(Score) <= 10$ $\And$ $Score$ does not prefix with “0”}
                \State S.append($int(Score)$)
            \EndIf
        \EndFor
        \For{$K_w$ in $K$}   
            \If{$K_w$ in $N$ $|$ $max(S) >= 5$} 
                \State \textbf{return} "Moderate-to-Severe Pain"
            \EndIf 
        \EndFor
        \State \textbf{return} "Mild Pain"
        \EndProcedure
    \end{algorithmic}
\end{algorithm}

For categorizing patients to mild and moderate-to-severe pain groups, we applied a rule-based approach. Patients whose pain scores exceeded 5 were classified as individuals experiencing a moderate-to-severe pain level, and others were placed in the mild pain group. The details of the pain scale-based algorithm are given in Algorithm \ref{alg:rule_base}.

\subsection{Approach \RNum{3}: Medication-based Approach}
\label{sec:medication_app1}

The medication records of 2,044 patients were extracted from the structured EMR data and stored in a separate CSV file in the format of the chart note data, with each record linked to the chart note by a unique patient ID. Each data row in the CSV file contains a patient ID and the list of medications prescribed to that patient for each physician encounter. Given the context of a patient having multiple records within the original dataset, we used the unique ID  assigned to each OA patient. By employing this method, we combined the medications from various records pertaining to each patient into a singular record. This combined record now comprises an extensive list of prescribed medications and alternative treatments. Then all patients' records from the 797 study sample were compiled into a multi-row dataset with each row representing a different patient ID with the corresponding medication information. The patient raw data was passed to the algorithm we developed for medication-based pain level categorization as listed in Algorithm \ref{alg:Medication_to_Pain_Level_al}. We also compiled a list of analgesics, i.e. all pain medications and treatments considered in this study were sent as the $2^{nd}$ parameter to the algorithm. The analgesics were categorized based on pain levels and compiled as a dictionary of medications labeled with pain levels. This was passed to the algorithm as the 3rd parameter. The algorithm \ref{alg:Medication_to_Pain_Level_al} delineates the procedural steps employed for extracting medication information from the structured notes and subsequently mapping them to pain levels. We first mapped pain levels to 797 patients in the study samples, and then we selected the same 156 patients in the final gold standard samples and evaluated the results.

\begin{algorithm}
\caption{Medication-based Approach} \label{alg:Medication_to_Pain_Level_al}
    \begin{algorithmic}[1]
        \Procedure{medication}{$PatientData$, $Analgesics$, $MedDict$}
        %\STATE Initialize one dictionary that contains analgesics as keys and chemical names as values: $D1$
        \State Initialize one empty dictionary: $D2$
        \State Initialize one empty dictionary: $D3$
        %\STATE $Patient\_med^{dict}$ = a dictionary containing the demographic numbers and medical information for each patient.
        %\STATE $Painlevel\_med^{dict}$ = a dictionary containing the pain levels and their corresponding medications.
        \For{$patientID, patientMedList$ in $PatientData$} 
            \For{$medication$ in $patientMedList$}
                \For{$analgesic, chemical$ in $Analgesics$}
                    \If{$medication$ in $analgesic$ or $medication$ in $chemical$)}
                        \State $D2[patientID]$.append($medication$)
                    \EndIf
                \EndFor
            \EndFor
            \For{$patientID, medication$ in $D2$}
                \For{$dictPainlevel, dictMed$ in $MedDict$}
                    \If{$medication$ in $dictMed$} 
                        \State $D3[patientID] = dictPainlevel$
                    \EndIf 
                \EndFor
            \EndFor
        \EndFor
        \State \textbf{return} D2, D3
        \EndProcedure
    \end{algorithmic}
\end{algorithm}

\begin{table}[h]
\caption{List of Analgesics Per Class.\label{tbl:Analgesics}}
 \footnotesize
\begin{tabularx}{\linewidth}{@{}@{}c*{6}{c}@{}}
\hline
\textbf{Class}          & \textbf{Chemical Name} 
\\ \hline
\textbf{ACETAMINOPHEN}  & ACETAMINOPHEN
\\ \hline
\textbf{TOPICAL NSAIDs} & DICLOFENAC
\\ \hline
\textbf{ORAL NSAID}     & \begin{tabular}[c]{@{}c@{}}ORAL NSAID, CELECOXIB, \\DICLOFENAL, IBUPROFEN, \\INDOMETHACIN, KETOPROFEN, \\ KETOROLAC, MEFENAMIC ACID, \\MELOXICAM, NAPROXEN, \\PIROXICAM, SULINDAC, \\TIAPROFENIC ACID\end{tabular} 
\\ \hline
\textbf{STRONG OPIOIDS} & \begin{tabular}[c]{@{}c@{}}CODEINE, BUPRENORPHINE, \\FENTANYL, OXYCODONE, \\HYDROCODONE, HYDROMORPHONE, \\MORPHINE,  METHADONE, \\TAPENTADOL\end{tabular}                                                       \\ \hline
\textbf{TRAMADOL}       & TRAMADOL
\\ \hline
\textbf{DULOXETINE}     & DULOXETINE
\\ \hline
\end{tabularx}
\end{table}

Next, we categorized the medications in the list of analgesics based on pain levels into a dictionary of medications. In accordance with the guidelines of WHO Analgesic Ladder\footnote{https://www.ncbi.nlm.nih.gov/books/NBK554435/}, all analgesic chemicals can be classified into three categories: mild, moderate, and severe pain. To categorize patients into mild and moderate-to-severe pain levels, we combined the moderate pain category with the severe pain category to establish the moderate-to-severe pain category as shown in Table \ref{tbl:medication}. 
\begin{enumerate}
  \item Mild pain: Non-opioid analgesics with or without adjuvants.
  \item Moderate Pain: Weak opioids (hydrocodone, codeine, tramadol) with or without non-opioid analgesics and with or without adjuvants.
  \item Severe and persistent Pain: Potent opioids (morphine, methadone, fentanyl, oxycodone, buprenorphine, tapentadol, hydromorphone, oxymorphone) with or without non-opioid analgesics, and with or without adjuvants.
\end{enumerate}

%we labeled the medications in Table \ref{medication} to mild and moderate-to-severe to indicate which medications are commonly prescribed to patients suffering from mild and moderate-to-severe pain level patients. This helped on improving the detection of pain levels for the patients.

\begin{table}[h]
\centering
\caption{Dictionary of Medications Categorized based on Pain Levels} 
 \label{tbl:medication}
 \footnotesize
\begin{tabularx}{\linewidth}{@{}@{}c*{6}{c}@{}}
\hline
\textbf{Pain Levels} & \textbf{Medications} \\ \hline
\textbf{Mild} & \begin{tabular}[c]{@{}c@{}}ACETAMINOPHEN, DICLOFENAC, \\ CELECOXIB,  DICLOFENAL, DULOXETINE\\ FLURBIPROFEN, IBUPROFEN,\\  INDOMETHACIN, KETOPROFEN,\\ KETOROLAC,  MEFENAMIC ACID, \\ MELOXICAM, NAPROXEN, PIROXICAM, \\ SULINDAC, TIAPROFENIC ACID\end{tabular} \\ \hline
\textbf{Moderate-to-Severe} & \begin{tabular}[c]{@{}c@{}}TRAMADOL, HYDROCODONE,\\ CODEINE,  BUPRENORPHINE,\\ FENTANYL, OXYCODONE\\ HYDROCODONE, HYDROMORPHONE,\\ MORPHINE,  METHADONE, TAPENTADOL\end{tabular} \\ \hline
\end{tabularx}
\end{table}

\textbf{Other Forms of Treatments:} The lists of medications prescribed to patients were for treating numerous health problems and not only pain. To extract and consider all pain related treatments, we formulated a dictionary of keywords specifically designed to discern OA-related surgical procedures, such as "knee surgery" and "hip surgery," along with OA-related mobility aids like "walker" and "cane", and referral-related information, such as "pain clinic".  %The detail of keywords are shown in Table \ref{tbl:other_form_treatments}. 
%Furthermore, we also incorporated keywords pertaining to pain management, including "pain clinic". 

%\begin{table}[h]
%\centering
%\caption{Keyword Dictionary for Other Forms of Treatments}  \label{tbl:other_form_treatments}
%\begin{tabular}{@{}ll@{}}
%\hline
%\multicolumn{1}{c}{\textbf{Category}} & \multicolumn{1}{c}{\textbf{Keyword}} \\\hline
%\textbf{OA Related Surgery}           & knee surgery, hip surgery            \\ \hline
%\textbf{Mobility Aid}                 & walker, cane, wheelchair             \\ \hline
%\textbf{Referral to Pain Clinic}      & pain clinic                          \\ \hline
%\end{tabular}
%\end{table}

%Patients whose list of medications had any mention of these keywords, were categorized under moderate-to-severe pain level as these treatments are generally prescribed for moderate or higher pain levels. Others were categorized under mild pain level. 

\textbf{Categorize Patients:} If any of the medications prescribed to a patient appeared under the category of moderate-to-severe pain level, or contained any of the keywords indicating other form of treatments for moderate-to-severe pain, then the patient was categorized under moderate-to-severe pain level, and otherwise under the mild pain level as shown in lines 11-17 in Algorithm \ref{alg:Medication_to_Pain_Level_al}.  %The ensuing algorithm (\ref{alg:Medication_to_Pain_Level_al}) delineates the procedural steps employed for extracting medication information from the structured notes and subsequently mapping them to pain levels.

\subsection{Integration}
\label{sec:integration}
Our system employs an integrated approach called PLeDO as shown in Figure \ref{fig:overall_flowchart} that combines the results of Approach \RNum{1}, \RNum{2}, and \RNum{3} to generalize across most chart notes, even when specific tools or scales are unavailable. While the pain scale-based systems are limited in their coverage of all possible tools and their mentions in chart notes, the integration of expression-based and medication-based methods ensures broader applicability. By leveraging this multi-method approach, we enhance the accuracy and generalizability of pain level detection across diverse chart notes. To combine the outcomes, we checked if any of the approaches indicated a moderate-to-severe pain level for a patient. If so, the patient was labeled as a moderate-to-severe pain level patient in the final assessment. Otherwise, the patient was labeled as a mild pain level patient.

%After obtaining the results of Approach \RNum{1}, \RNum{2}, and \RNum{3}, we integrated the 3 approaches to develop our integrated pain level detection tool for OA called PLeDO as shown in Fig. \ref{fig:overall_flowchart}. To combine the outcomes, we checked if any of the approaches indicated a moderate-to-severe pain level for a patient. If so, the patient was labeled as a moderate-to-severe pain level patient in the final assessment. Otherwise, the patient was labeled as a mild pain level patient.

\begin{table*}[]
\centering
\caption{Manual evaluation results and ablation study.} 
\begin{tabular}{@{}@{}l*{6}{c}@{}}
\hline
                                                                                                                                              & Accuracy       & Precision (PPV)      & Recall (Sensitivity)       & Specificity    & F1             & AUROC          \\ \hline
Pain Scale-based Approach                                                                                                                            & 0.551          & 0.572          & 0.589          & 0.589          & 0.536          & 0.589          \\ \hline
Medication-based Approach                                                                                                                                      & 0.551          & 0.560          & 0.575          & 0.575          & 0.532          & 0.575          \\ \hline
SPaDe                                                                                                                         & 0.577          & 0.541          & 0.549          & 0.549          & 0.533          & 0.549          \\ \hline
\begin{tabular}[c]{@{}l@{}}PLeDO\\   - Pain Scale-based Approach\\  + Medication-based Approach\\  + Pain Expression-based Approach\end{tabular} & 0.551          & 0.566          & 0.582          & 0.582          & 0.534          & 0.582          \\ \hline
\begin{tabular}[c]{@{}l@{}}PLeDO\\  + Pain Scale-based Approach\\  - Medication-based Approach\\  + SPaDe\end{tabular} & 0.622          & 0.547          & 0.552          & 0.552          & 0.548          & 0.551          \\ \hline
\begin{tabular}[c]{@{}l@{}}PLeDO\\    + Pain Scale-based Approach\\  + Medication-based Approach\\  - SPaDe\end{tabular} & \textbf{0.660} & 0.601          & 0.614          & 0.614          & 0.604          & 0.614          \\ \hline
\begin{tabular}[c]{@{}l@{}}PLeDO\\  + Pain Scale-based Approach\\  + Medication-based Approach\\  + SPaDe\end{tabular} & \textbf{0.660} & \textbf{0.605} & \textbf{0.621} & \textbf{0.621} & \textbf{0.608} & \textbf{0.621} \\ \hline
\multicolumn{4}{l}{\small * "+" indicates inclusion, and "-" indicates exclusion.} \\
\end{tabular} \label{tbl:table_result}
\end{table*}

\section{Validation and Results}
\label{sec:valres}

The results from experimental validation of our approaches in categorizing of the patients into mild and moderate-to-severe pain levels are discussed below. 

\subsection{Validation}
\label{sec:valid}

We used the gold standard data to validate our approaches. %Our approaches were run on the whole dataset of 797 patients which contained the gold standard data. 
Rather than limiting the application of our clustering method to the 156 gold validation datasets, we applied it across all 797 data points. This strategy ensures that, in instances of encountering unseen data, we can easily classify it into the appropriate category by leveraging our pre-trained centroids. The results from the unsupervised clustering approach of unlabeled data was validated using the manually labeled gold standard data. The same gold standard data was used to validate the medication-based and the pain scale based approaches. However, the manual evaluation did not consider the rigorous inspection of the medication and pain scale data. We report the validation of each approach in the results table. Evaluation metrics such as accuracy, precision (PPV), recall (sensitivity), specificity, F1 score, and AU-ROC were calculated for the gold standard validation dataset to report the performance.  
%, reserving the 156 datasets specifically for validation purposes

\subsection{Results}
\label{sec:results}

\begin{figure}[h]
    \centering
    \includegraphics[width=1\linewidth]{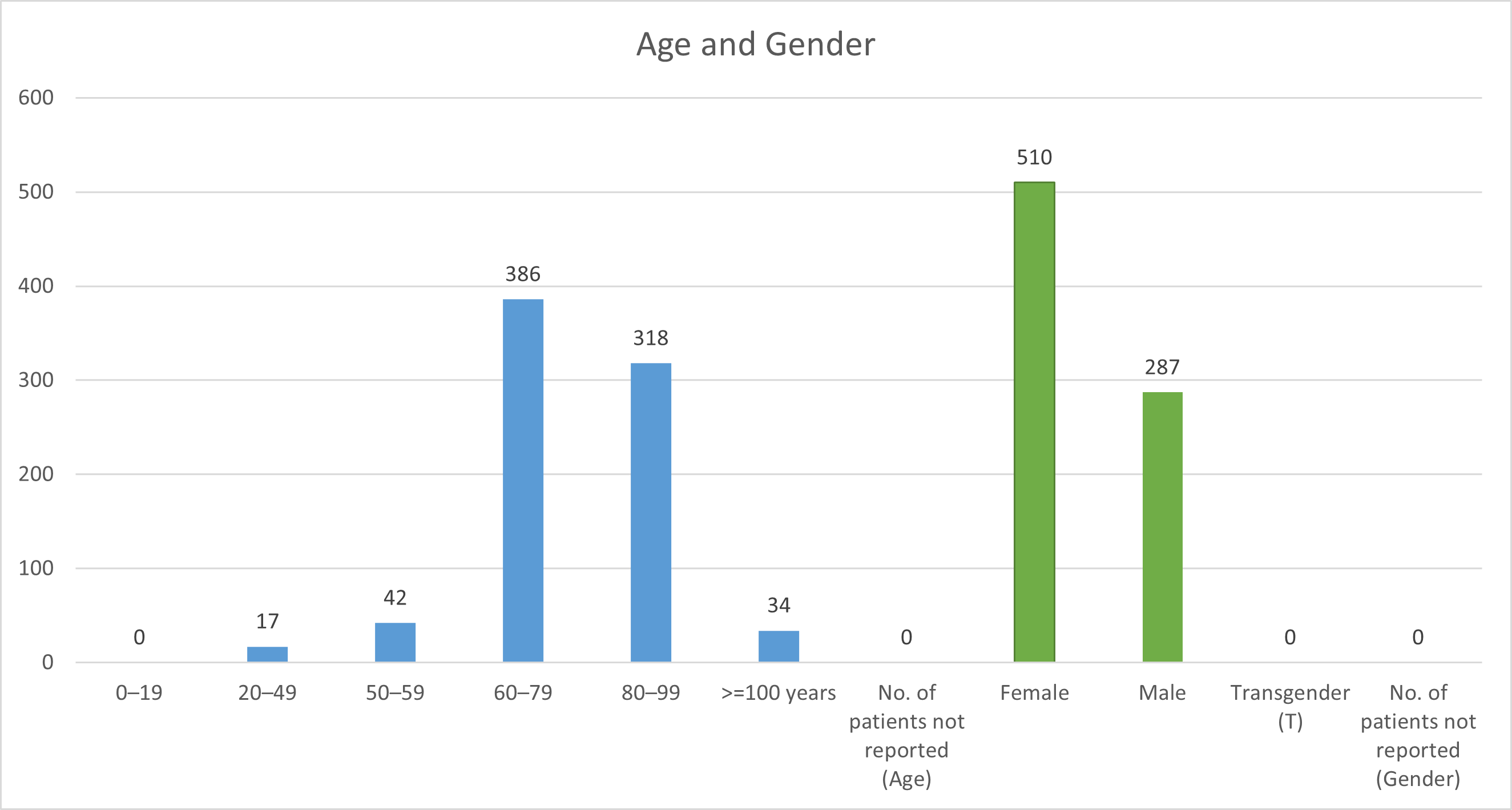}
    \caption{Gender and Age based Analysis of the Study Population.}
    \label{fig:GBA}
\end{figure}

We present the pain level categorization results in TABLE \ref{tbl:table_result}. A simple statistical analysis was also performed to see the gender and age distribution of the patients in our sample dataset with 797 patients as shown in Figure \ref{fig:GBA}. 

\subsection{Ablation Study}
\label{sec:ablation}

To provide an ablation study, we gradually combined multiple approaches as each approach was implemented as an independent parallel system with no dependency on the other systems. Finally, we combined all three approaches to present the results of our integrated approach PLeDO, which achieved the best performance. With additional information from the other approaches, the results improved a little. 

\subsection{Observations}
\label{sec:observation}

PLeDO achieved the best accuracy of 0.66 with an F1-score of 0.608 and an AU-ROC of 0.621. The ablation study demonstrates the significance of each component's inclusion on the final outcomes. Analysis of the population distribution shows the following trend (see Figure \ref{fig:GBA}).
\begin{itemize}
    \item There are more female patients with OA patients compared to male patients.
    \item Patients of age between 60-79 are more likely to have OA.
\end{itemize}

\subsection{Discussion }
\label{sec:discussion}

\subsubsection{Key Finding:} 

This study allowed us to explore the quality of EMR notes and treatment patterns for OA pain in primary care setting in Canada. The aim of this study was to see if pain expressions recorded in the chart notes are consistent with the treatment patterns. The following are the key takeaways from this study. Exploration of pain related expressions from the medical chart notes proved to be extremely challenging as it varies widely based on individual nature which is influenced by demography, culture, situation, and objective assessments for reporting. Our exploration of synonyms for mild, moderate, and severe pain using online tools as described in Section \ref{sec:unsupervised} returned words that were in some cases very distantly related to pain expressions. By changing the number of words in each of PLWDs, we can get different results which may be explored in future studies. Too few words will evade some of the expressions found in the chart notes and too many words will introduce confusion and noise in the process. We performed many iterations testing with different sets of words in the PLWDs before finalizing the lists of words. Also the expressions can change based on treatments as the pain conditions improve or deteriorate. We considered the highest level of pain expression for each patient's medical chart notes spanning multiple years. It is possible to categorize each note but not all notes have content indicating OA pain. Selecting good notes with pain information can introduce bias. 

Comorbidity in patients is another source of noise in the data. Pain can be the result of many different health problems and isolating OA related pain reports is a challenging task. Also, when other family members were mentioned or family history was narrated by the patient, it got more difficult to extract accurate information about the patient from the text data. 

We extracted paragraphs from the chart notes using OA related keywords but this list is not exhaustive. Any keyword related information extraction heavily relies on the set of keywords used for extracting the information and can either provide too much or too little information based on the number and quality of the keywords.

For the pain scale-related approach, we found that a standard practice was not followed in the data, and the scales were not always used to report pain levels. According to the results, its standalone performance was comparable to the medication-based approach but slightly lower than SPaDe. Therefore, if a standardized pain scale were consistently used, it may lead to improved and more reliable reporting of pain levels. Regarding medications, pain is a common problem for many diseases, and the medications are also general for all types of pain management. So, for OA specific pain, medications are not the best way to categorize the pain level. We considered all medications prescribed to the patients without filtering which medication was given for OA pain management. Therefore, we see that the results are not great. 

Regarding medications, pain is a common problem for many diseases and the medications are also general for all types of pain management. So, for OA specific pain, medications are not the best way to categorize the pain level. We considered all medications prescribed to the patients without filtering which medication was given for OA pain management. Therefore, we see that the results are not great. \\
\indent When comparing individual approaches, SPaDe demonstrates the highest accuracy but exhibits lower precision and recall. This outcome stems from SPaDe's strong performance in predicting the majority class (moderate-to-severe pain level), which constitutes a significant portion of the dataset (111 of 156 samples). However, accuracy alone does not provide insight into class-specific performance. A model can achieve high accuracy by excelling with the majority class while underperforming on the minority class (mild pain level, with only 45 samples). Lower precision suggests a higher rate of false positives, while lower recall (sensitivity) indicates a higher rate of false negatives. Interestingly, the ablation study shows that removing the pain scale-based approach results in the largest drop in accuracy, indicating its strong influence within the integrated framework. At the same time, removing SPaDe does not reduce overall accuracy in the combined setting. This highlights the strength of SPaDe, which, as an unsupervised clustering method, achieves substantial prediction consistency with both the scale-based (rule-based) approach and the medication-based approach that relies on human-defined rules and annotations. SPaDe’s generalizability is noteworthy. Unlike the pain scale-based approach, which depends on the presence of specific scales in chart notes, or the medication-based approach, which cannot cover all possible medications, SPaDe does not rely on such predefined elements. This makes it more flexible and adaptable to diverse datasets. \\
\indent Finally, the combination of all 3 approaches gives better results than any single approach. However, pain level detection is a difficult problem to address just using pain expressions from the unstructured data. Further exploration is needed to perhaps select a more specific population to focus on only pain management.

\subsubsection{Comparison with Existing Literature:}
Most previous research in this domain has focused on supervised learning using image data \cite{b2}\cite{b51}\cite{b59}, text data \cite{b47}\cite{b48}, or numerical data \cite{b45}\cite{b49}, often requiring extensive expert annotation. However, obtaining high-quality labels for medical data is both costly and time-consuming, which limits the scalability of such approaches. In recent years, large language models \cite{b9}\cite{b11}\cite{b19} have demonstrated exceptional performance across various domains. However, their propensity for hallucination and lack of interpretability present significant challenges for applications in sensitive areas such as the medical domain. Additionally, their substantial computational demands raise concerns about scalability and efficiency, further limiting their practicality in resource-constrained settings. This study was conducted in a secure setting without any internet access based on the established policies governing any work with real patients’ data. This limited our ability to download, fine-tune and examine some of the large language models. Furthermore, there were only few data points which were insufficient to train some of the large models.
We addressed the above limitations by developing an unsupervised clustering method, which generated predictions without relying on training labels. This approach enhances robustness, particularly when dealing with noisy data such as physicians' chart notes. Although clustering methods have been applied to pain level identification in prior research \cite{b49}, these studies only focus on numerical data, which is less complex and challenging compared to unstructured textual data. Additionally, we investigate the correlation between medication usage and pain levels to determine whether this relationship can further improve pain level identification. By combining structured medication data with unstructured text from chart notes, our system offers a comprehensive and innovative solution for pain identification in OA patients.

\subsubsection{Limitation:} 
This study is conducted using real-world clinical data from a single regional primary care network. As documentation practices, prescribing patterns, and patient characteristics may vary across regions, the findings may reflect patterns specific to this dataset and introduce regional bias in the results.
All EMR data used in this study were highly sensitive. For privacy and regulatory compliance, all experiments were conducted in a secure air-gapped environment with no external network access. This restriction limited the use of externally hosted large language models and certain high-compute methods, as such tools may introduce potential risks of sensitive information exposure.

\section{Conclusion}
\label{sec:conclusion}

Pain is a critical health problem that can greatly aggravate the quality of life of OA patients. Pain can have many subjective descriptions which make automated categorization of patients' chart notes based on pain severity a very challenging problem. Understanding the treatment pattern for different pain levels can lead to improved patient care and the discovery of new drugs. This study demonstrates the potential of using NLP and ML techniques in analyzing unstructured clinical data to extract valuable insights about pain severity among primary care patients diagnosed with OA. We developed 3 different approaches to pain level categorization using primary care unstructured chart note data and structured medication data. The main synonym based pain level detection approach, SPaDe, finds pain related expressions from chart notes using semantic similarity matching technique with the help of 3 pain level word dictionaries and then applies unsupervised clustering to group these expressions and the corresponding notes to mild and moderate-to-severe pain categories. The other two approaches, using medication and pain scale also provides similar accuracy. But our integrated approach, PLeDO, was able to achieve the best result.

In the future, we plan to investigate more focused osteoarthritis patient populations with fewer comorbidities and explore advanced clinical NLP tools such as cTAKES and MetaMap \cite{b39,b40} to improve concept extraction from unstructured EMR notes. As larger annotated datasets become available, supervised and semi-supervised approaches, including domain-specific transformer models such as MedicalBERT \cite{b41}, may be evaluated and compared with the proposed framework. We also plan to incorporate additional structured clinical variables, explore alternative clustering methods, and investigate privacy-preserving locally deployable foundation models. Finally, multimodal approaches that combine radiology images \cite{b42} with chart-note data may further improve osteoarthritis pain severity detection and characterization.

%In the future, we plan to explore more focused patient populations with less comorbidity to understand if such technique can perform better. We also plan to apply better NLP tagging tools customized for medical data such as cTAKES and MetaMap \cite{b39}\cite{b40}. We want to try other embedding methods such as the MedicalBERT \cite{b41}, and implement machine learning models by including more information from the structured data such as referral, age, gender to get a better understanding of the pain level. Another interesting path would be to combine the radiology images \cite{b42} with the chart note data to develop a more advanced approach to OA pain level detection and categorization of affected bone joints using multimodal data.

%\section*{Acknowledgment}
%We would like to express our deepest appreciation to Pfizer and Mitacs for funding this project.

%\section*{Statements and declarations}

%\subsection*{Consent to participate}
%Not applicable.

%\subsection*{Consent for publication}
%Not applicable.

%\subsection*{Declaration of conflicting interest}
%All authors have nothing to declare.

%\subsection*{Ethical considerations}
%The ethics approval number for this study is [details omitted for double-anonymized peer review]%6032120 SCOMP-010-21.

%\subsection*{Funding statement}
%[details omitted for double-anonymized peer review]

\end{document}